# Vision Transformers versus convolutional neural networks for fine-grained orchid genus identification in a species-rich, data-poor flora: a controlled benchmark on the Orchidaceae of New Guinea

Reza Saputra[1,2,*], Diah Harnoni Apriyanti[3], André Schuiteman[4], Kurt Metzger[5], Ashley Field[6,7], Katharina Nargar[7,8], William Edwards[1]

[1] College of Science and Engineering, James Cook University. McGregor Rd, Smithfield, Cairns, QLD 4878, Australia

[2] Southwest Papua Natural Resources Conservation Agency, Ministry of Forestry, Indonesia. Jalan Klamono KM 16, Sorong, Southwest Papua Province, Indonesia

[3] The Directorate of Scientific Collection Management, National Research and Innovation Agency (BRIN), Republic of Indonesia, Gedung Kehati, KST Soekarno BRIN, Jl. Raya Jakarta - Bogor KM 46, Cibinong, Kabupaten Bogor, Jawa Barat, 16911, Indonesia.

[4] Science Directorate, Royal Botanic Gardens, Kew, Richmond, TW9 3AB, UK

[5] Aiyura, Eastern Highlands Province, Papua New Guinea

[6] Queensland Herbarium, Department of the Environment, Tourism, Science and Innovation (DETSI), Mount Coot-tha Botanic Gardens, 4870, QLD, Australia

[7] Australian Tropical Herbarium, James Cook University, Building E1, McGregor Rd, Smithfield, Cairns, QLD 4878, Australia

[8] Australian National Herbarium, Commonwealth Industrial and Scientific Research Organisation (CSIRO), Clunies Ross Street, Canberra, 2601, ACT, Australia

* Correspondence: reza.saputra@jcu.edu.au

## Abstract

1. New Guinea is the world's richest island flora (~2,856 orchid species), yet most species are represented by only a handful of photographs—far fewer than direct species-level classification requires. Methods for fine-grained identification in such species-rich, data-poor floras using vision deep-learning models are needed, and it remains unclear which backbone architecture and pretraining strategy best support them.

2. We built a two-stage system that first predicts the *genus* of a query photograph and then retrieves visually similar reference images of candidate *species* using FAISS (Facebook AI Similarity Search). We ran a controlled comparison of four pretrained backbones—two Vision Transformers (ViTs; DINOv2, BioCLIP 2) and two convolutional neural networks (CNNs; ConvNeXt V2-L, EfficientNetV2-L)—fine-tuned under an identical protocol on a single fixed, species-stratified partition of 16,701 photographs spanning 120 genera and 1,350 species, and assessed accuracy, calibration (predicted-confidence reliability), error structure (confusion-matrix analysis), species retrieval, and open-set detection of novel genera.

3. DINOv2 attained the best genus performance (macro top-1 66.9%, 95% confidence interval, CI, 63.7–70.6; global top-1 88.9%); both Vision Transformers outranked both CNNs, with DINOv2's confidence interval disjoint from those of both CNNs, and general-purpose self-supervised pretraining (DINOv2) outperformed domain-matched biological pretraining (BioCLIP 2) by 7.1 points of macro top-1, although these two backbones also differed in input resolution (448 vs 224 px). Errors concentrated on two abundant genera that act as error attractors, drawing misclassifications from many smaller genera. DINOv2 embeddings achieved species Recall@5 of 86.6% and genus Recall@5 of 98.7%; post-hoc temperature scaling reduced every backbone's Expected Calibration Error to ≈0.03 without changing accuracy; and a distance-based open-set gate flagged genuinely unseen genera (mean area under the receiver-operating-characteristic curve, AUROC, 0.958).

4. A self-supervised Vision-Transformer backbone combined with embedding retrieval is an effective, deployable strategy for fine-grained identification in species-rich, data-poor floras. The system is released as an open web application (the New Guinea Orchid Identifier), offering a practical template for other hyperdiverse, under-documented taxa.

**Keywords:** biodiversity informatics, data scarcity, fine-grained image classification, image retrieval, New Guinea, Orchidaceae, self-supervised learning, Vision Transformer

# 1. Introduction

New Guinea is the most floristically diverse island on Earth, with an expert-verified checklist of 13,634 vascular plant species, 68% of them endemic (Cámara-Leret et al., 2020). The Orchidaceae are its single largest family: 2,856 species in roughly 133 genera have been documented (Cámara-Leret et al., 2020; Schuiteman, 1995; de Vogel et al., 2014), making the region a global hotspot of orchid evolutionary distinctiveness and a priority for conservation (Vitt et al., 2023). Identifying these orchids is difficult even for specialists—many genera are separated by subtle floral micro-characters—and expert capacity is scarce relative to the scale of undescribed and under-documented diversity. The difficulty is compounded by spatially uneven collecting: species-distribution modelling of New Guinean orchids was motivated by the need to infer richness and turnover from incomplete collection records, and emphasised strong geographic structure in both richness and assemblage composition (Vollering et al., 2016; Yudaputra et al., 2024). Taxonomic work continues to revise genera and describe new species, illustrating that the target flora is not a closed visual catalogue but an actively developing taxonomic system (Saputra et al., 2020, 2023; Kolanowska et al., 2021).

Automated, image-based identification has transformed botanical practice over the past decade (Borowiec et al., 2022). Citizen-science platforms such as Pl@ntNet now serve hundreds of millions of identifications using deep convolutional neural networks (CNNs; O'Shea & Nash, 2015; Affouard et al., 2017; Lefort et al., 2026), and purpose-built classifiers have been reported for orchids specifically, typically on small datasets of a few hundred to a few thousand images and a handful of species (Arwatchananukul et al., 2020; Apriyanti et al., 2023). However, these systems generally assume that each target class is represented by enough labelled images to train a direct species classifier. In a flora such as New Guinea's that assumption fails: the average species is represented by fewer than five photographs, a regime in which per-species classification is statistically infeasible and the class distribution is severely long-tailed (Van Horn et al., 2018; Cui et al., 2019). Automated plant recognition is now a mature computer-vision problem, but plant identification "in the wild" remains fine-grained because images vary in organ, viewpoint, background, phenology, and image quality, while many taxa differ by small inter-class visual cues (Wäldchen & Mäder, 2018; Šulc & Matas, 2017; Terry et al., 2020). Recent plant-recognition benchmarks have therefore compared CNNs, Vision Transformers (ViTs), and deep-embedding kNN systems at scale, reinforcing that architectural choices and retrieval formulation are central design decisions rather than implementation details (Picek et al., 2022).

Two further questions remain unsettled for fine-grained botanical recognition. First, *which backbone architecture* should anchor such a system? ViTs (Dosovitskiy et al., 2021) now rival or exceed CNNs on many benchmarks, and self-supervised Transformers such as DINOv2 (Oquab et al., 2024) produce general-purpose features without labels, while modern CNNs such as ConvNeXt V2 (Woo et al., 2023) and EfficientNetV2 (Tan & Le, 2021) remain strong and efficient. Second, does *domain-matched pretraining*—for example the biological vision–language model BioCLIP and its successor BioCLIP 2 (Stevens et al., 2024; Gu et al., 2025)—outperform general-purpose pretraining for a narrow taxonomic task? These questions are rarely addressed under controlled conditions on the same data.

Here we present a two-stage identification system for the orchids of New Guinea and use it as the setting for a controlled architecture benchmark. Our contributions are: (1) a two-stage pipeline (Figure 1) that combines genus-level classification with FAISS (Facebook AI Similarity Search) -based species retrieval (Johnson et al., 2021), designed explicitly for data-scarce floras; (2) a controlled comparison of two Vision-Transformer and two convolutional backbones, fine-tuned under an identical protocol on a single fixed, species-stratified partition, that disentangles the effect of architecture family from that of pretraining domain;

(3) a detailed characterisation of the best model, including per-genus accuracy, calibration, its long-tailed error pattern (dominant genera absorbing misclassifications in the confusion matrix), and how genera are organised in its embedding space; and (4) an open, deployed web application (the New Guinea Orchid Identifier or NGOIv3-DINOv2-L).

## 2. Materials and Methods

Figure 1 gives an overview of the two-stage system. Stage 1 predicts the genus of a query photograph with a fine-tuned Vision-Transformer backbone; Stage 2 restricts a FAISS similarity index to the predicted genus and retrieves the most similar reference photographs as species suggestions. An open-set distance gate between the two stages flags probable novel genera for expert review rather than forcing a confident misclassification.

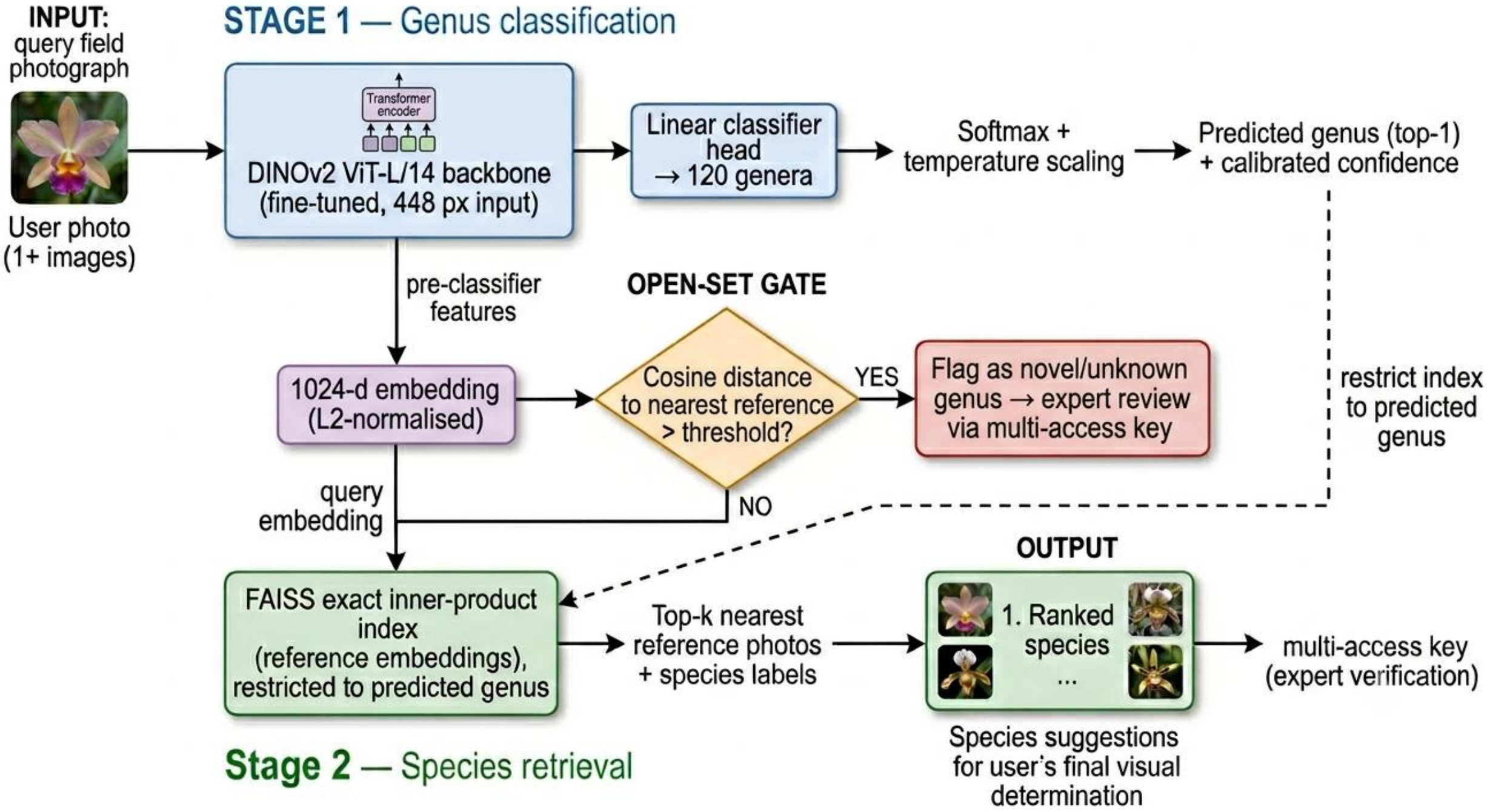

**Figure 1.** Overview of the two-stage orchid identification pipeline. In Stage 1 (genus classification), a DINOv2 ViT-L/14 backbone (fine-tuned, 448 px) feeds a linear classifier head over the 120 genera followed by softmax with post-hoc temperature scaling, yielding the predicted genus (top-1) and a calibrated confidence. The backbone's pre-classifier, L2-normalised 1024-dimensional embedding plays two roles: an open-set gate scores its cosine distance to the nearest known reference embedding—queries beyond a threshold are flagged as a probable novel/unknown genus and routed to expert review—while the same query embedding drives Stage 2. In Stage 2 (species retrieval), an exact FAISS inner-product index, restricted to the predicted genus (dashed arrow), returns the top-k nearest reference photographs and their species labels as ranked species suggestions for the user's final visual determination.

## 2.1 Dataset

We assembled an initial pool of 28,718 photographs of New Guinea orchids, drawn from curated public and contributed sources (orchidsnewguinea.com and associated collections, iNaturalist, and supplementary field photography). Because mixed botanical datasets frequently contain herbarium scans and botanical illustrations whose visual domain differs sharply from live photographs, an automated image-type screening step—a zero-shot CLIP (ViT-L/14; Radford et al., 2021) classifier that scored each image against ensembled text prompts for live, herbarium, and illustration classes—was applied, and its output was then curated by hand, with all borderline cases reviewed manually. After screening and curation, 16,701 live field photographs spanning 120 genera and 1,350 species were retained for training and evaluation; including herbarium specimens or line drawings would otherwise introduce domain-gap artefacts. The dataset has an extremely imbalanced class distribution. A few plant groups, like *Bulbophyllum* and *Dendrobium*, have hundreds of photos, while half of the genera have 21 or fewer.

## 2.2 Stratified fixed partition

A stratified split based on species labels was used to divide the dataset into fixed training, validation, and test sets (random seed 42). We carefully guarantee that photos from a specific species are grouped within a single partition instead of using a naive random image-level split. By preventing data leakage from specimen-specific features, this ensures the model learns genus-level morphology rather than memorising individual specimens. The final split comprises 11,677 training, 2,629 validation, and 2,395 test photographs. The 58 genera in the test split are the only ones that can be evaluated at the genus level due to the dataset's imbalanced-class distribution. For every experimental baseline, these partitions remained unchanged.

## 2.3 Stage 1 — genus classification

The genus classifier was selected empirically by comparing four pretrained backbones (Table 1): two Vision Transformers—DINOv2 ViT-L/14 (general self-supervised pretraining; Oquab et al., 2024) and BioCLIP 2 ViT-L/14 (hierarchical biological vision–language contrastive pretraining; Gu et al., 2025)—and two CNNs—ConvNeXt V2-L (fully-convolutional masked-autoencoder pretraining; Woo et al., 2023) and EfficientNetV2-L (supervised ImageNet-21k pretraining; Tan & Le, 2021). Each backbone was fitted with a single linear classification head over the 120 genera, applied to its pooled feature representation.

**Table 1.** The four pretrained backbones compared under an identical fine-tuning protocol.

| Backbone | Family | Pretraining | Input (px) | Embed. dim | Params (M) |
|---|---|---|---|---|---|
| DINOv2 ViT-L/14 | ViT | Self-supervised (natural images) | 448 | 1024 | 300 |
| BioCLIP 2 ViT-L/14 | ViT | Contrastive vision–language (biological) | 224 | 768 | 304 |
| ConvNeXt V2-L | CNN | Masked autoencoder (ImageNet) | 384 | 1536 | 198 |
| EfficientNetV2-L | CNN | Supervised (ImageNet-21k) | 448 | 1280 | 119 |

An identical training protocol was applied to all four backbones so the comparison is not confounded by training-regime differences. Each model was fine-tuned in two phases—a short warm-up (3 epochs) in which only the classification head is adapted, followed by end-to-end fine-tuning of the full network (up to 22 epochs, early-stopped with patience 8)—using cross-entropy loss with label smoothing ($\varepsilon = 0.1$) and no class-balanced resampling. Optimisation used AdamW (weight decay 0.05) with discriminative learning rates (head $1\times10^{-3}$, backbone $2\times10^{-5}$) under a warm-up-stable-decay schedule, and an exponential moving average of the weights (decay 0.9998) was tracked throughout. Crucially, the deployed checkpoint for each model was selected by best validation global top-1 accuracy rather than by validation macro-F1, which we found discarded better checkpoints. Each backbone was fine-tuned and evaluated at its standard operating resolution (Table 1), chosen per backbone to reflect the configuration in which it is conventionally fine-tuned and deployed rather than a single resolution imposed across all architectures; we therefore compare best-deployable configurations per backbone and return to the resulting resolution–architecture entanglement in Section 4. Models were implemented in PyTorch with the timm library and trained in the cloud (Google Colab) on a single NVIDIA L4 GPU (24 GB), using automatic mixed-precision (fp16 autocast) and multi-worker data loading; the batch size was fixed per model so that the optimisation trajectory was identical regardless of the run. Reference embeddings for Stage 2, species retrieval, were extracted on the same hardware under fp16 autocast.

## 2.4 Stage 2 — species retrieval

Because most species have too few images for direct classification, species-level suggestion is framed as visual retrieval. For the deployed backbone, the 1024-dimensional pooled feature vector (obtained from the pre-classifier representation) is extracted for every reference photograph, L2-normalised so that inner product equals cosine similarity, and indexed with FAISS using an exact inner-product index (Johnson et al., 2021). At query time, Stage 1 predicts the genus, the index is restricted to that genus, and the most similar reference

photographs—together with their species labels—are returned for the user to make a final visual determination. Multiple photographs of the same specimen can be fused by confidence-weighted averaging of their embeddings before retrieval. A coarse-to-fine design has precedent in plant recognition: Araújo et al. used botanical taxonomy to identify genus and species in a hierarchical strategy, motivated by the need to reduce dependence on many labelled examples and remain scalable to new plant species (Araújo et al., 2022).

## 2.5 Model testing and evaluation

Models were evaluated on the held-out test partition (Section 2.2), which contains no species seen during training, so reported genus performance reflects generalisation to unseen specimens rather than memorisation of specific plants. All four backbones were assessed on this identical test set, and the deployed backbone was additionally characterised in detail (per-genus accuracy, calibration (confidence reliability), error structure (confusion-matrix analysis), and the structure of the embedding space). Because the species-stratified partition leaves every test species without reference images, species retrieval was instead evaluated on a separate photo-level hold-out (described below). Finally, the ability of the system to flag genera it has never seen was tested with two complementary open-set protocols—an optimistic fixed-model leave-one-genus-out bound and a stricter leave-K-genera-out retraining test in which the withheld genera are genuinely unseen during training—with full results reported in Section 3.5. We also visualise the learned embedding space with a UMAP (Uniform Manifold Approximation and Projection) projection and quantify open-set separability using the area under the receiver-operating-characteristic curve (AUROC). The specific metrics and their uncertainty are defined below.

For genus classification we report global (micro-averaged) top-1 and top-5 accuracy and, as our primary metric, macro-averaged top-1 accuracy, which weights every genus equally and is therefore sensitive to performance on the long tail. Probabilistic calibration is quantified by the Expected Calibration Error (ECE; Guo et al., 2017), reported both as trained and after post-hoc temperature scaling—a single scalar temperature T fitted on the validation split by minimising the negative log-likelihood (Guo et al., 2017); because T divides all logits equally it cannot change the ranking, so top-1 and macro accuracy are invariant and only calibration is affected. Retrieval cannot be evaluated on the species-stratified partition—every test species has, by construction, no reference photographs—so it is assessed on a separate photo-level hold-out: for every species with at least two photographs, one photograph (fixed seed) is withheld as a query and the remainder form the reference index, and we report species- and genus-level Recall@k. Every reported metric is accompanied by a 95% confidence interval (CI) obtained by non-parametric bootstrap of the test set (1,000 resamples; Efron & Tibshirani, 1994).

# 3. Results

## 3.1 Backbone comparison: Vision Transformers versus CNNs

On the fixed test partition (2,395 photographs; 58 genera), the self-supervised Vision Transformer DINOv2 attained the best performance on every classification metric (Table 2, Figure 2), with macro top-1 accuracy of 66.9% (95% CI 63.7–70.6) and global top-1 of 88.9%. It was followed by the second Vision Transformer, BioCLIP 2 (macro top-1 59.8%, CI 56.2–63.1), and then the two CNNs, ConvNeXt V2-L (52.8%, CI 49.8–56.7) and EfficientNetV2-L (47.8%, CI 44.6–51.5). The architecture families separate cleanly: both Vision Transformers outranked both CNNs, and DINOv2's confidence interval is disjoint from those of both convolutional models. The same ordering held for global top-1, macro top-5, and macro top-1 restricted to genera with at least five test photographs (DINOv2 69.6%, BioCLIP 2 60.4%, ConvNeXt V2-L 58.0%, EfficientNetV2-L 52.1%).

**Table 2.** Backbone comparison on the fixed test partition (n = 2,395; 58 genera). Values are percentages except ECE. Macro top-1 is the primary metric; ECE (raw) is the as-trained Expected Calibration Error and ECE (post-TS) is after a single temperature scalar fitted on the validation split (Guo et al., 2017), which leaves top-1/accuracy unchanged. Best per column in bold in the text.

| Backbone | Family | Global T1 | Macro T1 | Macro T1 (≥5) | Macro T5 | ECE (raw) | ECE (post-TS) | Species R@5 |
|---|---|---|---|---|---|---|---|---|
| DINOv2 ViT-L/14 | ViT | **88.9** | **66.9** | **69.6** | **83.6** | 0.178 | 0.029 | **86.6** |
| BioCLIP 2 ViT-L/14 | ViT | 84.4 | 59.8 | 60.4 | 79.9 | 0.205 | 0.030 | 86.3 |
| ConvNeXt V2-L | CNN | 83.2 | 52.8 | 58.0 | 74.4 | 0.146 | 0.031 | 85.0 |
| EfficientNetV2-L | CNN | 80.5 | 47.8 | 52.1 | 72.2 | **0.142** | **0.028** | 84.2 |

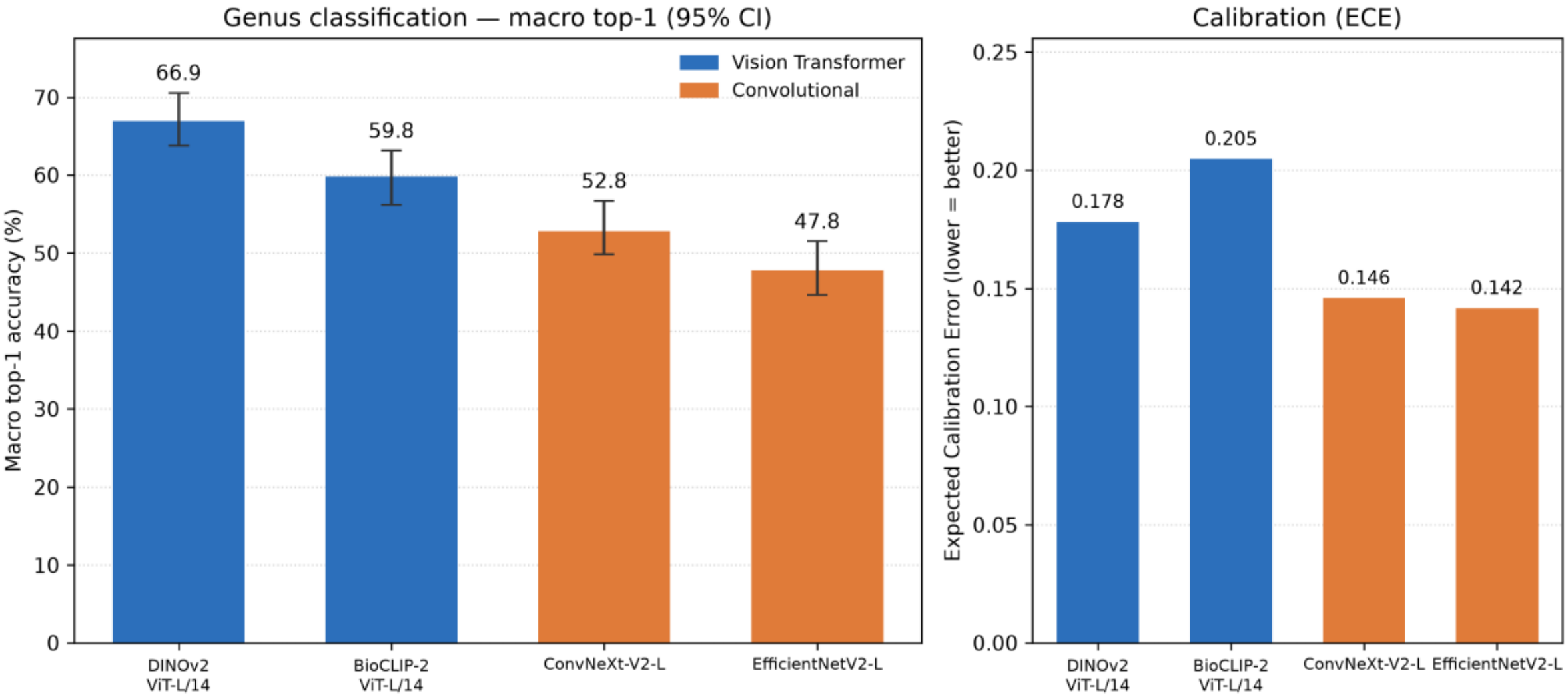

**Figure 2.** Genus-classification macro top-1 accuracy with 95% bootstrap confidence intervals (left) and Expected Calibration Error (right) for the four backbones, coloured by architecture family. Both Vision Transformers outrank both CNNs on accuracy, whereas the CNNs have lower as-trained calibration error; all four backbones reach ECE ≈ 0.03 after post-hoc temperature scaling (Figure 3, Table 2).

Two results stand out. First, the general-purpose self-supervised Transformer (DINOv2) outperformed the domain-matched biological model (BioCLIP 2) by 7.1 points of macro top-1. Because the two backbones share the ViT-L/14 architecture but were evaluated at different input resolutions (448 vs 224 px), this gap reflects the combined effect of pretraining strategy and input resolution rather than pretraining alone (Section 4); as deployed, the general-purpose self-supervised representation nonetheless transferred better to this fine-grained genus task than vision–language contrastive pretraining on a broad biological corpus. Second, in their as-trained outputs calibration ran counter to accuracy: the CNNs were the best calibrated (ECE 0.142–0.146) and the Vision Transformers the worst (DINOv2 0.178, BioCLIP 2 0.205). This miscalibration was a systematic under-confidence—across the confidence range the empirical accuracy exceeded the stated confidence, so the reliability curve lies above the diagonal (Figure 3)—as expected when a 120-way softmax distributes probability mass over many genera. A single post-hoc temperature fitted on the validation split ($T < 1$, which sharpens the distribution; Guo et al., 2017) corrected this for every backbone, reducing ECE to ≈0.03 (Table 2, ECE post-TS) and erasing the Vision Transformers' calibration disadvantage; because the temperature divides all logits equally it is a monotonic rescaling, so top-1 and macro accuracy are unchanged. With macro top-1 as the primary criterion, DINOv2 was selected as the deployed backbone for Stage 1; because it also gave the best species retrieval of any backbone (Section 3.4), a single backbone serves both stages of the deployed system, and it is released as the NGOIv3-DINOv2-L checkpoint.

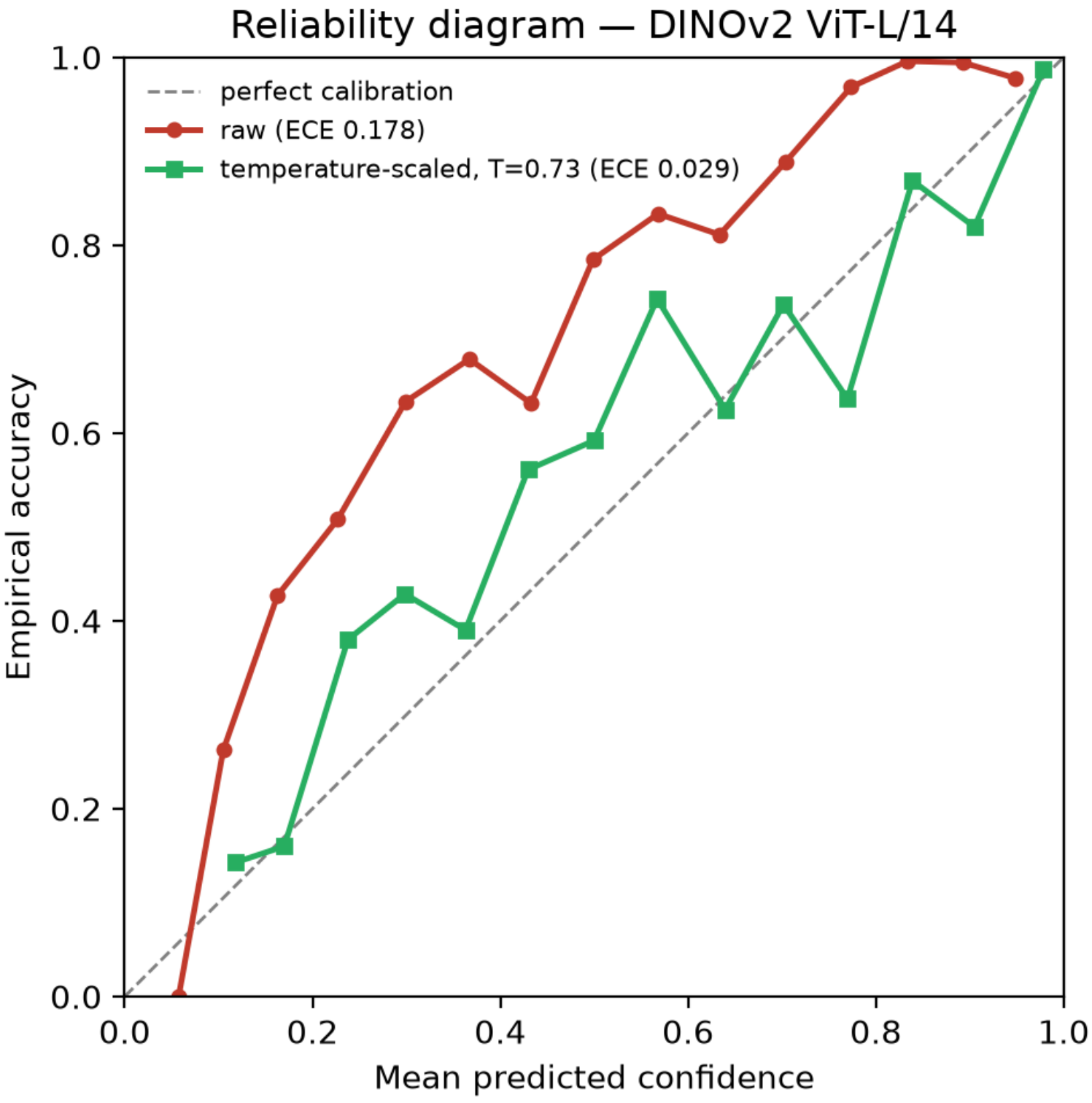

**Figure 3.** Reliability diagram for the deployed DINOv2 classifier on the test partition (15 equal-width confidence bins). The as-trained curve (red) lies above the diagonal across the confidence range—empirical accuracy exceeds the stated confidence—showing the classifier is systematically under-confident, a consequence of the 120-way softmax spreading probability mass across many genera. A single validation-fitted temperature (T = 0.73) sharpens the predictions onto the diagonal, lowering ECE from 0.178 to 0.029 without changing any prediction (top-1 accuracy is identical). Calibration differences across modern architectures are not settled by architecture family alone: recent large-scale analyses found that some newer non-convolutional models are among the best calibrated, while architecture, scale, pretraining, and distribution shift all interact (Minderer et al., 2021). The under-confidence observed here is therefore best presented as an empirical property of these fine-tuned orchid classifiers, with post-hoc temperature scaling as a practical correction that changes confidence without changing the top-1 ranking (Minderer et al., 2021; Ao et al., 2023).

## 3.2 Best model in detail — DINOv2 ViT-L/14

Performance of the deployed model was strongly support-dependent (Figure 4). The abundant genera were near-saturated (*Bulbophyllum*, n = 808: 98.9%; *Dendrobium*, n = 622: 97.9%), while the long tail of small genera drove the gap between global (88.9%) and macro (66.9%) accuracy. This pattern—high micro accuracy with a substantial macro penalty—is the expected signature of a severely long-tailed task and motivates reporting macro accuracy as the primary metric.

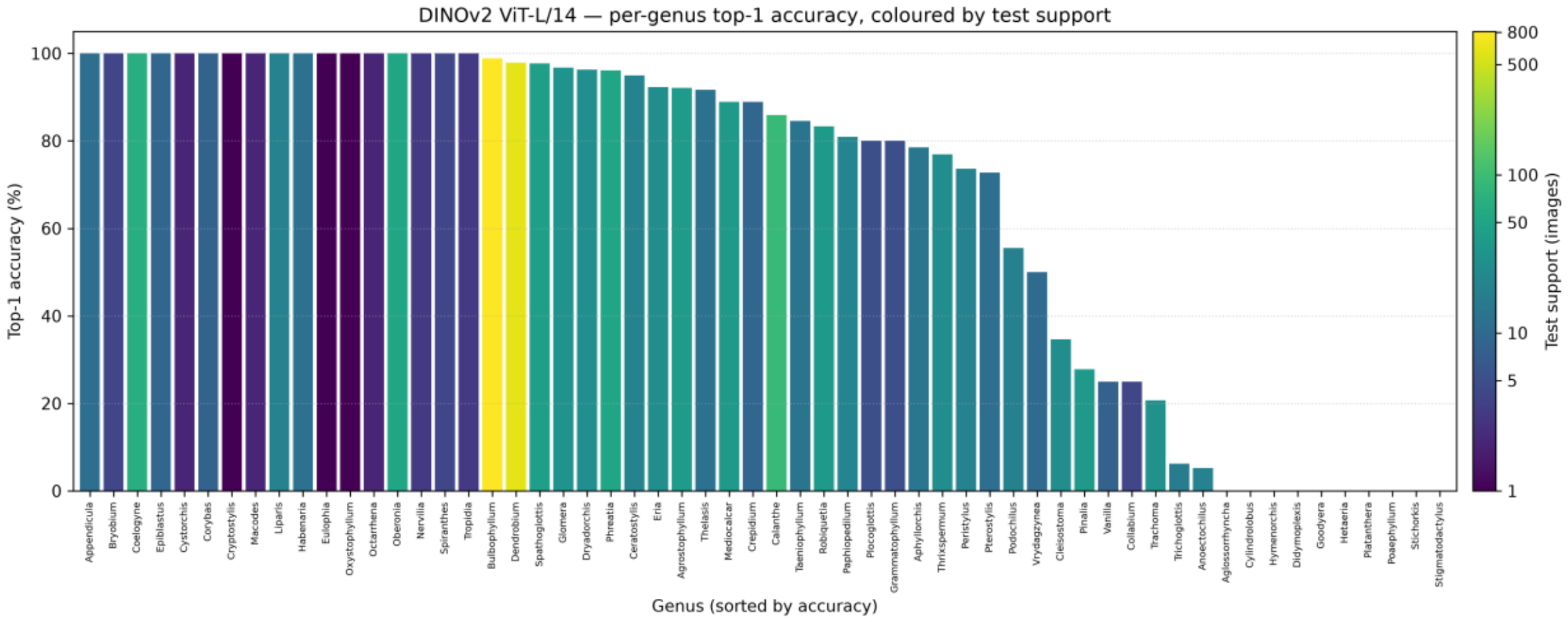

**Figure 4.** Per-genus top-1 accuracy of DINOv2 on the test partition, sorted by accuracy, with bars coloured by test support (number of images, log scale). Accuracy declines sharply for low-support genera in the long tail.

### 3.3 Confusion-matrix analysis: dominant genera as error attractors

The confusion structure was shared across all four backbones and dominated by two abundant genera that act as error attractors, which draw misclassifications from many smaller genera (Figure 5). Summed over the four models, misclassifications were absorbed overwhelmingly by *Dendrobium* (355 test images) and *Bulbophyllum* (186), the two largest classes. Several low-support genera collapsed to 0% top-1 in every model (*Aglossorrhyncha*, *Cylindrolobus*), being predicted as one of these dominant genera, and *Pinalia* → *Dendrobium* was the single largest confused pair (about 69% of *Pinalia* test images). DINOv2's advantage was concentrated precisely on the hardest genera (Table 3). It alone resisted the *Bulbophyllum* attractor for *Paphiopedilum*, reaching 81.0% top-1 versus ≤9.5% for the other three backbones, which lost 76–95% of their *Paphiopedilum* images to *Bulbophyllum*. Specifically, ConvNeXt V2-L, BioCLIP 2, and EfficientNetV2-L sent 16/21, 18/21, and 20/21 *Paphiopedilum* images to *Bulbophyllum*, respectively, versus only 4/21 for DINOv2.

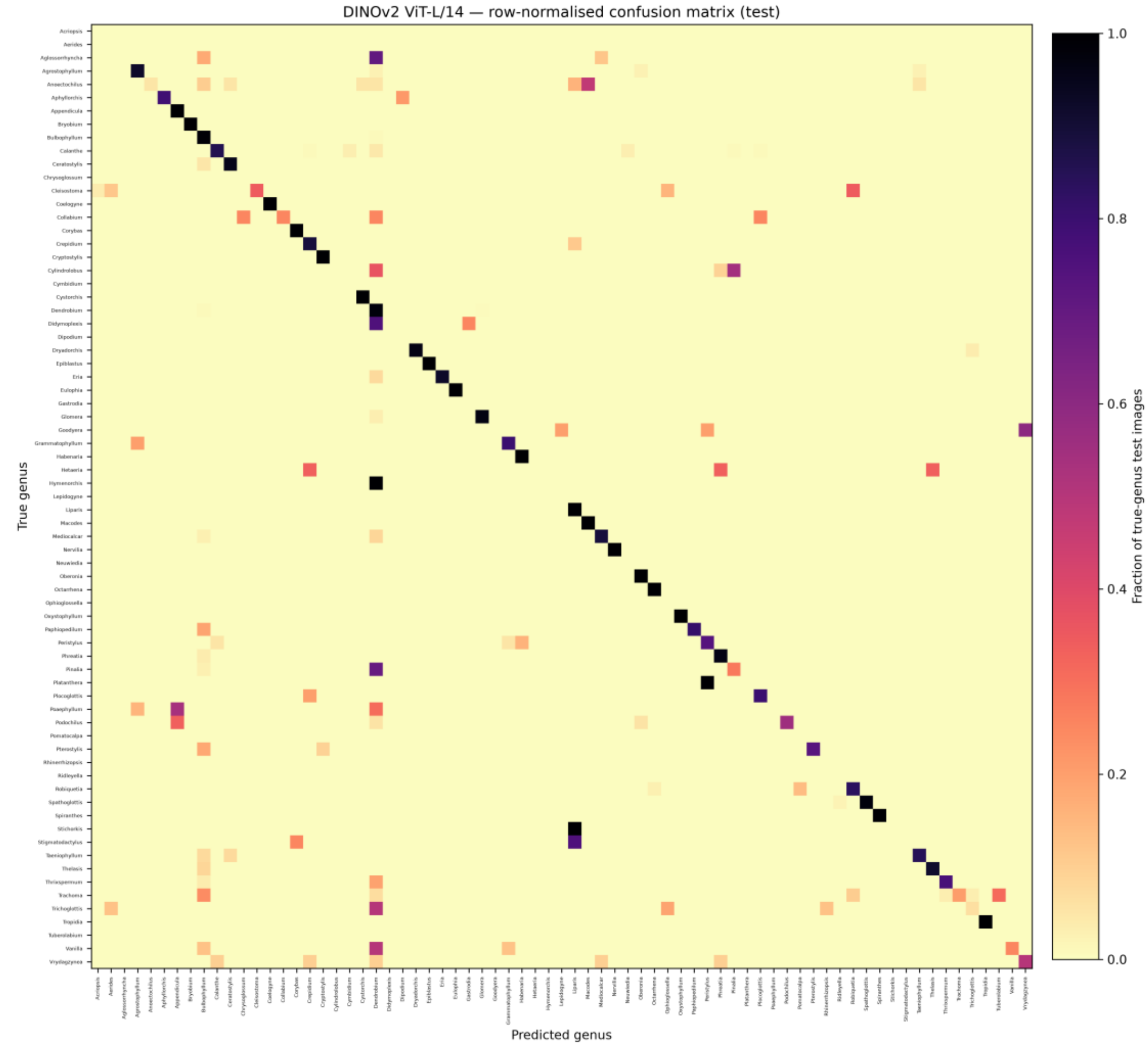

**Figure 5.** Row-normalised confusion matrix for DINOv2 on the test partition. The strong diagonal indicates correct classification; vertical bands at *Dendrobium* and *Bulbophyllum* reveal their role as error attractors for low-support genera.

**Table 3.** The ten hardest genera (lowest mean top-1 across models; test support n ≥ 10). Values are top-1 accuracy (%).

| Genus | n | Mean | DINOv2 | BioCLIP 2 | ConvNeXt V2-L | EfficientNetV2-L |
|---|---|---|---|---|---|---|
| *Aglossorrhyncha* | 17 | 0.0 | 0.0 | 0.0 | 0.0 | 0.0 |
| *Cylindrolobus* | 11 | 0.0 | 0.0 | 0.0 | 0.0 | 0.0 |
| *Anoectochilus* | 19 | 1.3 | 5.3 | 0.0 | 0.0 | 0.0 |
| *Trichoglottis* | 16 | 1.6 | 6.2 | 0.0 | 0.0 | 0.0 |
| *Poaephyllum* | 13 | 3.9 | 0.0 | 15.4 | 0.0 | 0.0 |
| *Trachoma* | 29 | 10.3 | 20.7 | 10.3 | 6.9 | 3.4 |
| *Pinalia* | 36 | 18.1 | 27.8 | 19.4 | 16.7 | 8.3 |
| *Vrydagzynea* | 10 | 20.0 | 50.0 | 20.0 | 10.0 | 0.0 |
| *Paphiopedilum* | 21 | 22.6 | 81.0 | 0.0 | 9.5 | 0.0 |
| *Cleisostoma* | 26 | 33.6 | 34.6 | 11.5 | 57.7 | 30.8 |

### 3.4 Species retrieval and intra-/inter-class variance

Using DINOv2 embeddings, the retrieval stage attained species Recall@5 of 86.6% (95% CI 84.7–88.4) and genus Recall@5 of 98.7% on the photo-level hold-out; the species figure narrowly exceeded that of every other backbone, so a single backbone wins both stages. A two-dimensional UMAP (Uniform Manifold Approximation and Projection) projection of the DINOv2 reference embeddings reveals that the representation is strongly genus-structured without any retrieval-stage supervision: the two most abundant genera occupy large, well-separated regions and smaller genera form compact, distinct islands (this geometry is displayed in Figure 7a for the leave-12-genera-out model of Section 3.5, whose embedding space is essentially identical to the deployed model's). This geometric separation is what makes nearest-neighbour retrieval viable for species suggestion in the absence of a species-level classifier; the diffuseness of the *Dendrobium* manifold mirrors its exceptional morphological breadth and its role as the dominant error sink in Stage 1. Strikingly, the embedding also resolves structure below the genus level. The two visually separate *Dendrobium* sub-clusters in Figure 7a correspond to a major infrageneric division: a compact island of the montane, bird-pollinated section *Calyptrochilus*—comprising 94% of all section-*Calyptrochilus* photographs together with allied montane species such as *D. cuthbertsonii*—set apart from the main mass of larger-plant sections (*Grastidium*, *Latouria*, *Spatulata*). That section-level morphology emerges without any taxonomic supervision reinforces why embedding retrieval is effective for within-genus species suggestion; section assignments follow the orchidsnewguinea.com database and remain subject to expert revision. The retrieval stage is also aligned with recent plant-recognition evidence that kNN in learned deep embeddings can outperform purely classifier-based formulations on large plant datasets, especially when the recognition problem is naturally open-ended and visually fine-grained (Picek et al., 2022).

### 3.5 Open-set detection of novel genera

The deployed model recognises 120 genera, but New Guinea harbours more, so a query may belong to a genus the system has never seen. We therefore tested whether such inputs can be flagged rather than silently misclassified, scoring each query by the cosine distance from its embedding to the nearest reference embedding—a larger distance indicating a more probable novel genus (Hendrycks & Gimpel, 2017; Vaze et al., 2022). We assessed this in two complementary ways: an optimistic *fixed-model* bound and a stricter *retraining* test. In biodiversity monitoring, open-set recognition is not a peripheral edge case: real deployments routinely encounter undescribed, unsampled, or geographically novel taxa absent from the training set (Villon et al., 2022; Chen et al., 2025).

In the first, with the deployed DINOv2 model held fixed, we performed leave-one-genus-out over every genus with at least ten photographs (88 genera): each genus in turn was removed from the reference bank and its photographs scored as unknowns. Genera proved strongly separable in the retrieval embedding (mean AUROC 0.963, median 0.977, range 0.736–0.999; 99% of genera above 0.80; Figure 6). This is an *optimistic* bound, however, because the backbone had been trained on all 120 genera and may therefore carry latent structure for the "withheld" genus.

To obtain a deployment-realistic estimate we ran a stricter leave-K-genera-out retraining test in which the withheld genera are genuinely unseen during training. Twelve genera spanning the full support range (16–274 photographs; deliberately including the morphologically distinctive slipper orchid *Paphiopedilum*) were removed from the training and validation sets, and DINOv2 was re-trained from the same initialisation under the identical protocol on the remaining 108 genera. The retrained backbone was essentially unchanged in closed-set quality (validation global top-1 85.1%), confirming that withholding rare genera did not degrade the representation. All 16,701 photographs—including the twelve unseen genera as queries—were then embedded with this hold-out model and each scored by cosine distance to its nearest *known* reference. The genuinely-unseen genera remained highly separable: mean per-genus AUROC 0.958 (median 0.966, range 0.894–0.997; pooled 0.961), with *Paphiopedilum* the hardest (0.894) and *Apostasia* the easiest (0.997).

A genus-by-genus comparison against the optimistic fixed-model bound (Table 4) makes this robustness concrete. Averaged over the same twelve genera, the strict estimate (0.958) was, if anything, marginally higher than the fixed-model bound (0.955; mean change +0.003, mean absolute change 0.018), and for five of the twelve genera open-set detection actually improved when the genus was genuinely unseen; only *Paphiopedilum* showed a substantial decline (−0.049). That the strict, retrained estimate matches—rather than collapses below—the fixed-model bound (pooled 0.961 vs 0.963) indicates that the separability reflects genuine representational structure of the DINOv2 embedding rather than memorised training genera. Two caveats temper a literal reading of the per-genus differences. First, the two protocols are not strictly identical: the fixed-model test scores a single genus as unknown against the remaining 119 known genera, whereas the strict test scores all twelve withheld genera as unknown against only 108 known genera, so the reference set and the in-distribution pool differ and the comparison is practical rather than perfectly controlled. Second, per-genus AUROC for the rarest genera carries appreciable sampling uncertainty. The near-equality nonetheless holds across both the pooled estimate and the paired per-genus values. Figure 7a shows the held-out *Paphiopedilum* photographs forming a compact, peripheral island in the embedding space, and Figure 7b that they sit at markedly larger distances from the known set than in-distribution photographs. As a concrete deployment operating point, we fix the threshold τ at the 95th percentile of the known-genus nearest-neighbour distances (τ = 0.29 in cosine distance, a 5% false-positive rate on known genera), at which 70% of genuinely novel-genus photographs are flagged; relaxing the tolerance to a 10% false-positive rate (τ = 0.21) raises this to 94%. In practice this lets the system abstain on probable novel genera and route them to expert review through the Lucid key, converting a silent error into an explicit "unknown—needs verification" outcome.

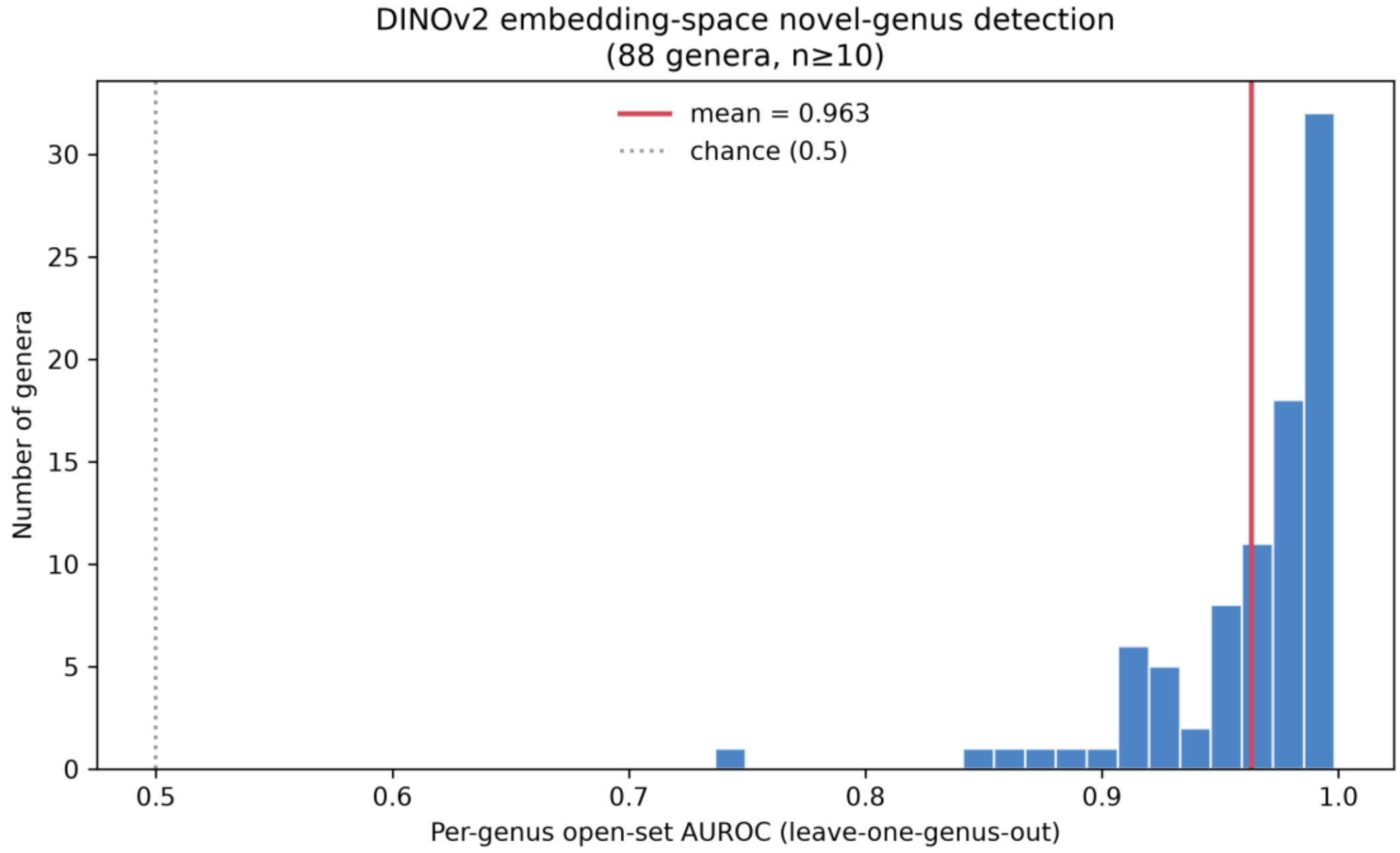

**Figure 6.** Open-set detection of novel genera by nearest-reference embedding distance (leave-one-genus-out, fixed DINOv2 model). Distribution of per-genus AUROC across 88 genera (n ≥ 10); mean 0.963. Most genera are highly separable from the known set.

**Table 4.** Paired open-set AUROC for the twelve held-out genera under both protocols: the optimistic fixed-model bound (leave-one-genus-out; DINOv2 trained on all 120 genera, the genus removed only from the reference bank) versus the strict retraining test (DINOv2 retrained on the remaining 108 genera, the genus genuinely unseen during training). Δ is strict minus fixed; positive values mean the genus was detected as novel at least as well when genuinely unseen. The two protocols are not strictly identical (fixed: one unknown genus vs 119 known; strict: twelve unknown vs 108 known), so the comparison is practical rather than perfectly controlled. Mean over the twelve genera: fixed 0.955, strict 0.958. Here n is the total number of photographs of each genus used as unseen queries, distinct from the test-partition support n reported in Table 3. Δ is computed from unrounded AUROC values and may differ by 0.001 from the difference of the three-decimal values shown.

| Held-out genus | n | Fixed-model AUROC | Strict AUROC | Δ (strict − fixed) |
|---|---|---|---|---|
| *Apostasia* | 16 | 0.998 | 0.997 | −0.002 |
| *Pterostylis* | 47 | 0.971 | 0.980 | +0.009 |
| *Chrysoglossum* | 21 | 0.981 | 0.975 | −0.006 |
| *Aglossorrhyncha* | 26 | 0.974 | 0.970 | −0.004 |

| *Oberonia* | 274 | 0.987 | 0.970 | −0.018 |
|---|---|---|---|---|
| *Robiquetia* | 172 | 0.952 | 0.966 | +0.014 |
| *Thrixspermum* | 75 | 0.979 | 0.966 | −0.014 |
| *Crepidium* | 105 | 0.963 | 0.962 | −0.001 |
| *Spiranthes* | 21 | 0.912 | 0.955 | +0.044 |
| *Dryadorchis* | 37 | 0.927 | 0.938 | +0.011 |
| *Vanda* | 44 | 0.878 | 0.927 | +0.049 |
| *Paphiopedilum* | 51 | 0.944 | 0.894 | −0.049 |
| Mean (12 genera) | — | 0.955 | 0.958 | +0.003 |

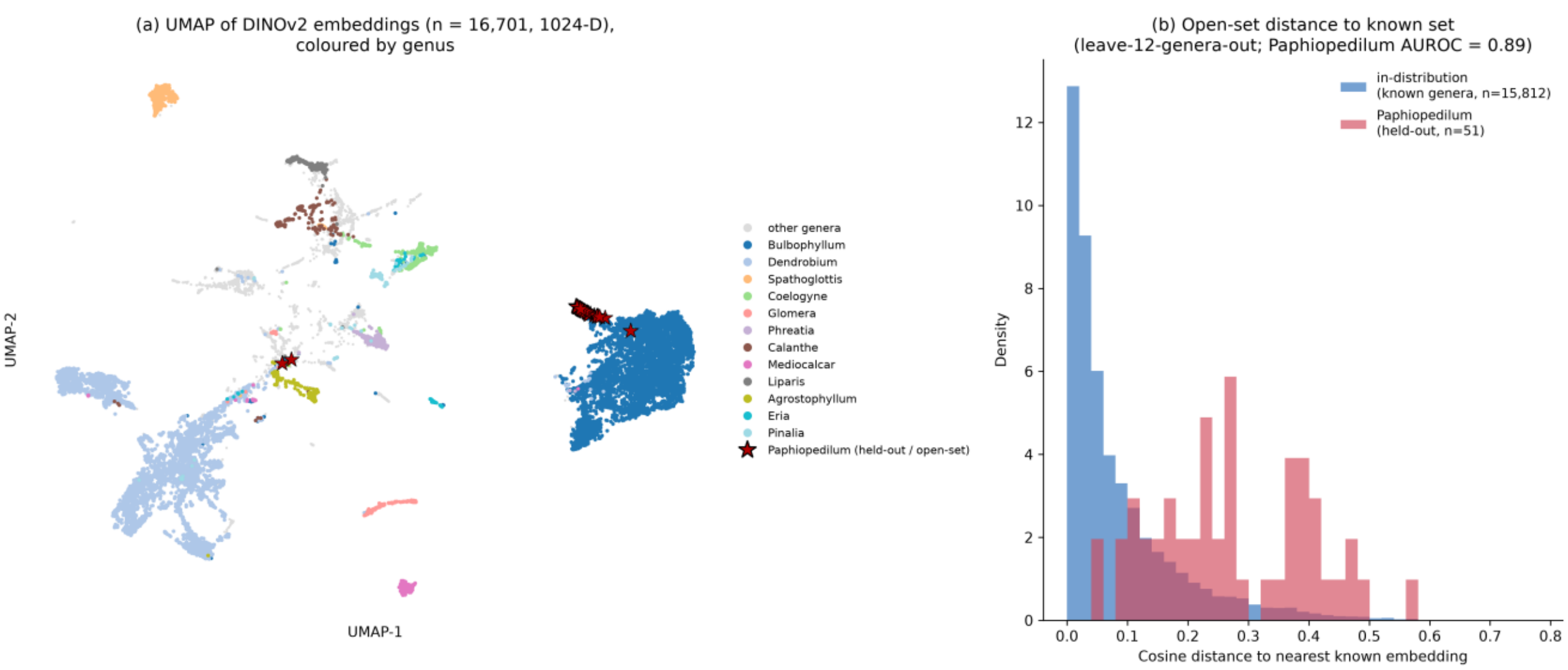

**Figure 7.** Embedding-space structure and open-set separation under the strict leave-12-genera-out retraining test. (a) UMAP projection (cosine metric) of the 16,701 DINOv2 embeddings from the hold-out model, with the twelve most photographed known genera coloured, remaining genera in grey, and the unseen exemplar *Paphiopedilum* overlaid as red stars; the unseen genus forms a compact peripheral island. (b) Cosine distance to the nearest known reference embedding for in-distribution (known) photographs versus the unseen *Paphiopedilum*, which separates from the known set with AUROC 0.89.

## 4. Discussion

Under identical settings with rigorous train-test separation, Vision Transformers clearly outperformed CNNs for fine-grained orchid genus recognition: both Transformers ranked above both convolutional models on the primary metric, with confidence intervals that do not overlap between the best Transformer and either convolutional network. This is consistent with the view that the global receptive field of self-attention (Dosovitskiy et al., 2021) is well-suited to capturing the spatially distributed floral micro-characters that distinguish orchid genera. Crucially, this advantage is not an artefact of input resolution: the lowest-resolution backbone in the comparison was itself a Vision Transformer (BioCLIP 2 at 224 px) yet outranked both CNNs running at equal or higher resolution, and at matched resolution (DINOv2 and EfficientNetV2-L both at 448 px) the Vision Transformer led by 19 points of macro top-1 (Section 4, Limitations). Fine-grained visual classification is a natural setting for testing Transformers because discriminative evidence may be distributed across small object parts rather than concentrated in a single local patch; for example, TransFG explicitly uses transformer attention to select discriminative patches and model their relations (He et al., 2022). More generally, representation analyses show that ViTs aggregate global information earlier than ResNets, offering a plausible mechanism for their advantage when global floral configuration and local detail jointly matter (Raghu et al., 2021).

More surprising is that general-purpose self-supervised pretraining outperformed domain-matched biological pretraining. DINOv2, trained on curated natural images without labels (Oquab et al., 2024), exceeded BioCLIP 2 (Gu et al., 2025) by 7.1 points of macro top-1, despite the latter's biological vision–language training. This comparison must be read with caution, however, because the two backbones were evaluated at different input resolutions (448 vs 224 px); pretraining strategy and resolution are therefore entangled, and the gap cannot be attributed to pretraining alone until a resolution-matched test is run (Section 4). Should the effect persist under matched resolution, a plausible explanation is that contrastive vision–language objectives optimise for cross-modal alignment and broad taxonomic coverage rather than for the within-family visual discrimination required here, whereas DINOv2's dense self-supervised features retain finer local detail. This echoes the broader finding that strong self-supervised features transfer remarkably well across domains, and cautions against assuming that domain-matched foundation models are always preferable for narrow downstream tasks. This interpretation is consistent with large-scale transfer studies showing that strong self-supervised visual features can outperform supervised baselines across many downstream tasks, while still finding that no single self-supervised method dominates all settings (Ericsson et al., 2021). It is also consistent with DINO-family evidence that self-supervised ViTs can yield spatially meaningful features and strong nearest-neighbour classifiers without labels, properties directly relevant to both genus classification and embedding retrieval here (Caron et al., 2021).

The error analysis exposes a long-tailed head-class-bias failure mode of practical importance: misclassifications concentrate on the two most abundant genera, and several rare genera are never recovered. This is the expected behaviour of cross-entropy training under extreme class imbalance (Cui et al., 2019) and suggests that targeted interventions—class-balanced objectives, additional reference imagery for the most absorbed genera, or hierarchical decision rules—could yield further gains beyond the choice of backbone. Encouragingly, the best backbone is also the most robust on the hardest genera, indicating that representation quality, not merely class frequency, governs tail performance. In the long-tailed-recognition literature, this pattern is usually described as head-class bias or majority class dominance:

abundant head classes dominate feature and classifier learning, while rare tail classes are disproportionately predicted as visually plausible head classes (Zhang et al., 2023; Xu et al., 2022). This confusion structure is therefore a useful biological description of the benchmark, but it is best interpreted as an instance of a broader head-to-tail bias rather than as an orchid-specific pathology.

The two-stage design directly addresses data scarcity. Rather than attempting infeasible 1,350-way species classification, the system predicts the more learnable genus and defers species determination to embedding retrieval and, ultimately, to the user. The strong genus structure of the DINOv2 embedding space (Figure 7a) and the high genus-level Recall@5 (98.7%) make this division of labour effective. In deployment, the system returns ranked species suggestions for the user's final visual determination, combining the throughput of machine vision with user/expert review and species information page. Robustness to the inevitable arrival of unrecognised genera is provided by the open-set distance score (Section 3.5): because that score remained discriminative even for genera entirely unseen during training (strict mean AUROC 0.958, essentially matching the fixed-model bound), the system can abstain on probable novel genera and defer them to expert review rather than misclassify them silently. This deployment choice is consistent with conservation-informatics work arguing that AI and citizen science are complementary when machine predictions accelerate data triage but expert validation preserves data quality (McClure et al., 2020). In biodiversity platforms, automated detection of uncertain or likely erroneous identifications has been proposed specifically to focus limited expert effort on the records most needing review (Saoud et al., 2020).

The infrageneric structure recovered within *Dendrobium* (Figure 7a) is biologically interpretable and indicates that the embedding encodes functional floral morphology rather than incidental image cues. The compact sub-cluster corresponds almost entirely to section *Calyptrochilus*, which is diagnosed by a distinctive floral architecture: a long, narrow, spur-like mentum held parallel to the ovary that lends the flower an elongate, tubular outline, and a simple lip whose sharply incurved, hood-like apex bears a fringed (fimbriate) margin that veils the column, the feature from which the section takes its name (de Vogel et al., 2014). These flowers are also vividly and persistently coloured (scarlet, orange, red, violet or yellow, and long-lived), a classic ornithophilous syndrome; birds are documented pollinators in *Calyptrochilus* and its close relatives *Pedilonum* and *Oxyglossum* (Burzacka-Hinz et al., 2025). This pollinator-driven floral gestalt explains why the section detaches as a discrete island: the few non-*Calyptrochilus* images that fall within it are predominantly *Pedilonum* (e.g., *D. secundum*, *D. hasseltii*), members of the same clade that share the same brightly coloured, tubular bird-flower habit, so the apparent leakage reflects genuine morphological convergence rather than classifier error. The contrast is informative because floral structure across *Dendrobium* is otherwise comparatively uniform, with relatively few qualitative differences and most variation residing in size and colour (Adams, 2011); against this background the divergent, ornithophilous floral form of *Calyptrochilus* is essentially the only morphological block distinct enough to crystallise without taxonomic supervision. That a section-level signal grounded in pollination ecology emerges without supervision reinforces the biological validity of the learned representation and, in turn, the suitability of embedding retrieval for within-genus species suggestion.

**Limitations.** First, each backbone was evaluated at its standard operating resolution (224–448 px; Table 1), so input resolution co-varies with backbone. This does not explain the architecture result: the lowest-resolution model in the comparison was itself a Vision Transformer (BioCLIP 2 at 224 px), yet it outranked both CNNs running at equal or higher resolution (ConvNeXt V2-L at 384 px, EfficientNetV2-L at 448 px), and at matched resolution (DINOv2 and EfficientNetV2-L both at 448 px) the Vision Transformer led by 19 points of macro top-1. The four backbones also differ in capacity and embedding dimension (Table 1; EfficientNetV2-L has 119 M parameters versus roughly 300 M for the Vision Transformers), so model size co-varies with architecture family and is not fully isolated by our protocol; that a larger CNN (ConvNeXt V2-L, 198 M) also trails both Transformers argues against capacity alone explaining the ordering, but a parameter-matched comparison remains future work. Resolution is, however, entangled with the pretraining comparison: DINOv2 and BioCLIP 2 share the ViT-L/14 architecture but were evaluated at 448 and 224 px respectively, so the 7.1-point gap between them reflects the combined effect of pretraining strategy and input resolution; a resolution-matched ablation (both backbones at 224 px) is required to isolate the pretraining contribution and is left for future work. Second, in their as-trained outputs the Vision Transformers were less well calibrated (more under-confident) than the CNNs; post-hoc temperature scaling reduced every backbone's ECE to ≈0.03 (Section 3.1), so this is readily corrected, but the temperature should be re-fitted if the model or input distribution changes. Third, retrieval quality is bounded by the coverage and label accuracy of the reference set; species absent from the reference index cannot be retrieved. Finally, the evaluation covers the 58 genera represented in a single fixed test partition; broader geographic and taxonomic validation is desirable. All reported confidence intervals are non-parametric bootstrap resamples of this single fixed test partition and therefore reflect test-set sampling but not training stochasticity, as each backbone was trained once from a single seed; repeated runs across seeds and splits would further probe the robustness of the architecture ranking. The Transformer advantage should not be read as universal: CNN locality and weight-sharing are useful inductive biases in small-data regimes, and hybrid ViT designs have been proposed specifically to recover sample efficiency through convolution-like initialization (d'Ascoli et al., 2021). The present benchmark therefore supports DINOv2 for this dataset and protocol, but does not rule out CNN or hybrid advantages under different resolution, augmentation, or pretraining regimes.

## 5. Conclusion

For fine-grained orchid identification in a species-rich, data-poor flora, a self-supervised Vision-Transformer backbone (DINOv2) combined with embedding-based species retrieval is an effective and deployable strategy. In a controlled benchmark, Vision Transformers outperformed CNNs; general-purpose self-supervised pretraining also outperformed domain-matched biological pretraining. However, this comparison used each backbone's standard input resolution rather than a resolution-matched setting. The resulting two-stage system—released as an open web application—offers a practical template for automated identification in other hyperdiverse, under-documented taxa.

## Data and Code Availability

The interactive web application, the New Guinea Orchid Identifier (genus classification and species retrieval), is openly deployed on Hugging Face Spaces (https://huggingface.co/spaces/Rezamonium/ong-orchid-identifierv3); the training and evaluation code and the deployed model weights are available in the repository linked therein. The associated project website is available at https://birdsheadorchid.id. The curated photographic dataset is described in detail above; reference imagery originates from orchidsnewguinea.com, iNaturalist, and contributed field collections. The raw images are subject to third-party licensing and cannot be redistributed in bulk; they remain available from their original sources subject to the originators' terms. The code and the best-performing model checkpoint of each backbone are archived on Zenodo (https://doi.org/10.5281/zenodo.21090831). The deployed backbone checkpoint is named NGOIv3-DINOv2-L.

## Acknowledgements

We thank the contributors and curators of orchidsnewguinea.com, the community of observers and contributors of iNaturalist, and the field photographers whose images made this work possible; and Jeffrey Champion for contributing photographs used in this study. This work was supported by the Australian Orchid Foundation.

## Author Contributions

RS: Conceptualization, Methodology, Software, Formal analysis, Writing – original draft. KM: Resources, Data curation. AF, AS, DHA and KN: Writing – review & editing. WE: Supervision. All authors read and approved the final manuscript.

## Conflict of Interest

The authors declare no conflict of interest.